\documentclass[pdflatex,sn-mathphys-num]{sn-jnl}

\usepackage{graphicx}%
\usepackage{multirow}%
\usepackage{amsmath,amssymb,amsfonts}%
\usepackage{amsthm}%
\usepackage{mathrsfs}%
\usepackage[title]{appendix}%
\usepackage{xcolor}%
\usepackage{textcomp}%
\usepackage{manyfoot}%
\usepackage{booktabs}%
\usepackage{algorithm}%
\usepackage{algorithmicx}%
\usepackage{algpseudocode}%
\usepackage{listings}%
\usepackage{float}
\usepackage[utf8]{inputenc} 
\usepackage[T1]{fontenc}    

\theoremstyle{thmstyleone}%
\theoremstyle{thmstyletwo}%

\theoremstyle{thmstylethree}%

\begin{document}

\title[Article Title]{Learning Fourier shapes to probe the geometric world of deep neural networks}


\author[1]{\fnm{Jian} \sur{Wang}}\email{wj851329121@stu.xjtu.edu.cn}

\author[1]{\fnm{Yixing} \sur{Yong}}\email{yongyx@stu.xjtu.edu.cn}

\author[1]{\fnm{Haixia} \sur{Bi}}\email{haixia.bi@mail.xjtu.edu.cn}

\author[1]{\fnm{Lijun} \sur{He}}\email{lijunhe@mail.xjtu.edu.cn}

\author*[1]{\fnm{Fan} \sur{Li}}\email{lifan@mail.xjtu.edu.cn}

\affil[1]{\orgdiv{Shaanxi Key Laboratory of Deep Space Exploration Intelligent Information Technology}, \orgname{School of Information and Communications Engineering}, 
	\orgaddress{\street{Xi’an Jiaotong University}, 
		\city{Xi’an}, \postcode{710049}, 
		\country{China}}}


\abstract{While both shape and texture are fundamental to visual recognition, research on deep neural networks (DNNs) has predominantly focused on the latter, leaving their geometric understanding poorly probed. Here, we show: first, that optimized shapes can act as potent semantic carriers, generating high-confidence classifications from inputs defined purely by their geometry; second, that they are high-fidelity interpretability tools that precisely isolate a model's salient regions; and third, that they constitute a new, generalizable adversarial paradigm capable of deceiving downstream visual tasks. This is achieved through an end-to-end differentiable framework that unifies a powerful Fourier series to parameterize arbitrary shapes, a winding number-based mapping to translate them into the pixel grid required by DNNs, and signal energy constraints that enhance optimization efficiency while ensuring physically plausible shapes. Our work provides a versatile framework for probing the geometric world of DNNs and opens new frontiers for challenging and understanding machine perception.}




\maketitle
\section{Introduction}
\begin{figure}[!t]
	\centering
	\includegraphics[width=1.0\textwidth]{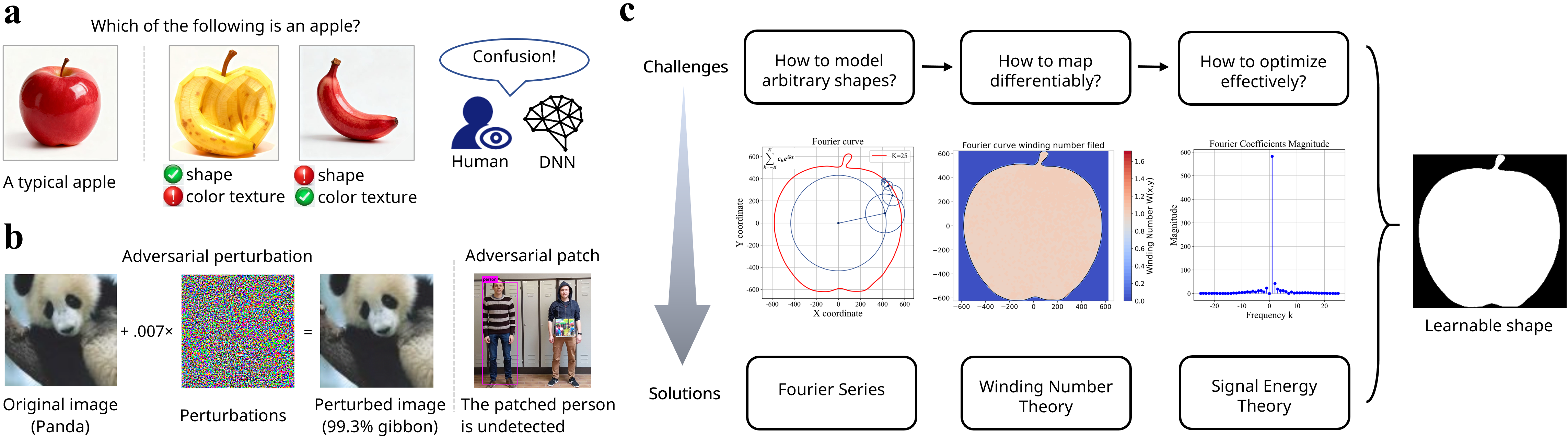}
	\caption{\textbf{Conceptual overview of adversarial shape learning}. \textbf{a}, Human and machine visual systems rely on consistent shape and appearance attributes for robust object recognition. When these attributes are mismatched, such as an apple's shape with a banana's texture, perceptual conflict arises, illustrating that shape is an independently salient attribute. \textbf{b}, Prior work on adversarial attacks primarily targets the appearance domain. This involves either adding subtle, global pixel perturbations to misclassify an image (e.g., a \emph{panda} recognized as a \emph{gibbon}) or deploying localized adversarial patches to cause detection failures. These methods operate on pixel values without explicitly manipulating underlying geometry. \textbf{c}, Our framework enables end-to-end differentiable optimization of object shapes for adversarial machine learning. It addresses three key challenges: (1) Shape parameterization: Arbitrary closed contours are represented by a compact set of Fourier series coefficients. (2) Differentiable mapping: A module based on the winding number theorem translates these coefficients into a 2D grid image, creating a differentiable bridge to DNNs. (3) Effective optimization: Regularization, inspired by signal energy theory, guides the learning process to ensure physically plausible shapes by constraining high-frequency components. This integrated pipeline allows for the discovery and optimization of effective adversarial shapes.
	Images in \textbf{b} are from ref. \cite{Goodfellow_Explaining_2014} and \cite{Thys_CVPRW2019}}\label{fig1}
\end{figure}

The remarkable ability of the human visual system \cite{Mahner_NMI2025,Doerig_NMI2025} to recognize objects relies on a sophisticated synthesis of two fundamental attributes: geometric shape and surface texture. Shape provides the structural scaffold of an object, defining its boundaries and identity, while texture and colour furnish the finer details of its appearance.  A significant distortion in either of these attributes can disrupt perception (Fig. \ref{fig1}a), suggesting that both are fundamental to robust recognition. A well-trained DNN should ideally mirror this biological duality \cite{Geirhos_ICLR2018}, leveraging both shape and texture to make robust inferences. However, the vast body of research exploring model vulnerabilities through adversarial attacks \cite{Szegedy_Intriguing2013,Goodfellow_Explaining_2014} has overwhelmingly focused on manipulating the texture domain \cite{Dutta_ICCV2025,Lee_CVPR2025,Fang_ICCV2025}. By searching for subtle perturbations in the high-dimensional pixel space, these methods have revealed profound weaknesses in modern DNNs, yet they have largely overlooked the equally fundamental axis of shape.

This intense focus on pixel-level manipulations (Fig. \ref{fig1}b), while fruitful for revealing model weaknesses, carries inherent limitations. Adversarial perturbations \cite{Ghaffari_NC2022,Veerabadran_NC2023,Paniagua_ICML2025}, typically composed of high-frequency signals, are largely confined to the digital domain and lack direct physical-world applicability. While physically realizable methods like adversarial patches \cite{Brown_adv_shape, Wang_TIP2025,Wang_TCSVT2025,Wang_Exoploring_ICCV2025} have been developed, they still operate by manipulating texture within a predefined boundary rather than the object's intrinsic geometry. Furthermore, the explanatory power \cite{Woods_NMI2019,Ignatiev_nips2019} of such pixel-based methods is often limited, as the resulting patterns lack clear semantic meaning for human observers \cite{Ilyas_2019NeurIPS,Simonyan2013,Zeiler_Visualizing}. This raises a critical question: can we move beyond the pixel grid to engage directly with a model's understanding of geometry? Exploring the domain of shape offers a path to creating more physically robust and interpretable methods for analyzing and challenging machine perception.

Directly optimizing an object's shape, however, presents a formidable technical challenge. A shape is a continuous, geometric entity typically described by abstract parameters, unlike the discrete grid of pixels a DNN processes. Bridging this gap for gradient-based optimization requires two key components: a shape representation that is expressive enough to describe a diverse family of forms, and a differentiable mapping to translate those parameters into a pixel array. An effective representation must therefore combine expressive power with optimization efficiency, providing a rich search space in which to discover effective adversarial geometries.
Existing approaches \cite{chen_deform_eccv2022,wei_inf_ijcv2024,wei_phy_cvpr2023,zhu_infrared_cvpr2022,zhu_folling_aaai2021,wei_TPAMI2023,wei_unifined_ICCV2023} have struggled to satisfy these requirements simultaneously. Methods that model shapes on a discrete grid \cite{wei_inf_ijcv2024,wei_phy_cvpr2023,zhu_infrared_cvpr2022,zhu_folling_aaai2021}, for instance, are differentiable but require complex, hand-crafted aggregation constraints to maintain coherence, which restricts the search space and scales poorly with resolution. Conversely, approaches using continuous spline-based representations \cite{wei_TPAMI2023,wei_unifined_ICCV2023} often lack a differentiable mapping, forcing a reliance on inefficient black-box optimization that yields poor scalability and performance.

Here, we propose a complete framework for learning adversarial shapes through a parametric Fourier series representation. Inspired by how complex signals can be decomposed into a sum of simple sinusoids, we model any arbitrary 2D closed contour using a compact set of Fourier coefficients. This representation allows us to generate a vast and intricate space of shapes by controlling the amplitude and phase of different frequency components. To bridge the gap between these abstract parameters and the pixel domain, we employ a differentiable mapping based on the winding number theorem from complex analysis, which analytically \emph{draws} the shape defined by the Fourier coefficients onto a 2D grid, generating an image where each pixel's value is a function of its location relative to the contour. The entire pipeline, from Fourier coefficients to a rasterized shape image, is fully end-to-end differentiable. Furthermore, by introducing regularization constraints based on signal energy principles, we guide the optimization towards generating shapes that are both physically plausible and adversarially potent.

This framework allows us to explore the role of shape in machine perception with unprecedented control. Our experiments reveal three key findings. \textbf{First, we demonstrate that shape alone is a powerful carrier of semantic information}, capable of generating high-confidence classifications from a DNN even in the complete absence of texture; moreover, the strength of this semantic information gracefully scales with the shape's complexity via the number of Fourier terms. \textbf{Second, we repurpose our method as a high-fidelity interpretability tool}. For a given input image, by optimizing a shape mask to be as small as possible while preserving correct classification, we can isolate a model's region of interest with sharper, more interpretable boundaries than existing methods like Grad-CAM \cite{Grad-cam}. Conversely, we show that occluding a small but critical region, while retaining the vast majority of the original image, is sufficient to guarantee misclassification. \textbf{Finally, we establish adversarial shapes as a generalizable attack paradigm}, analogous to colour-based adversarial patches with fixed shapes. We show that by optimizing the shape of a patch while keeping its colour fixed, we can effectively cause a person covered by the shape to evade the state-of-the-art object detectors, demonstrating the method's applicability to diverse downstream vision tasks.

\begin{figure}[!t]
	\centering
	\includegraphics[width=1.0\textwidth]{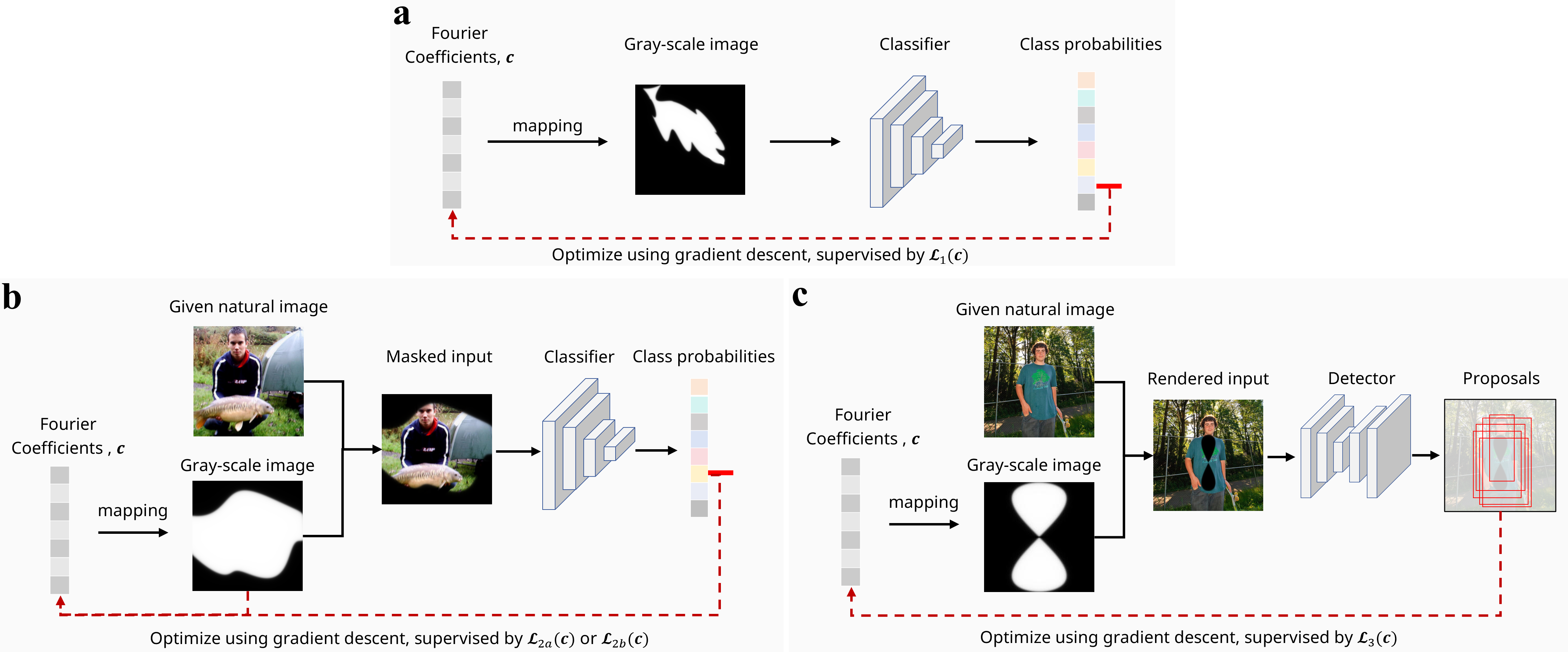}
	\caption{\textbf{Overview of the three experimental frameworks enabled by the differentiable shape learning pipeline}. \textbf{a}, Experiment 1: Class-specific shape generation. A set of Fourier coefficients, $\mathbf{c} = \{c_k\}_{k=-K}^{K}$, is converted via the differentiable mapping into a gray-scale shape image. This image is fed directly into a classifier. The coefficients are optimized using gradient descent to maximize the classification confidence for a chosen target class, demonstrating the semantic representation capability of shape alone. \textbf{b}, Experiment 2: Shape as an interpretability tool. The Fourier coefficients are mapped to a gray-scale image, which is used as a mask on a given natural image. The masked input is fed into a classifier. The coefficients are optimized using two symmetric objectives: (1) to maximize the confidence for the true class while simultaneously minimizing the shape's area, thereby isolating the model's minimal salient region; or (2) to minimize the true class confidence while maximizing the shape's area, identifying the minimal critical region whose occlusion causes misclassification. \textbf{c}, Experiment 3: Shape as a generalizable adversarial paradigm. The Fourier coefficients are mapped to a gray-scale image, which is then rendered as an occlusion patch onto a target (e.g., a person) in a natural image. The rendered input is fed into an object detector. The coefficients are optimized to minimize the detection confidence scores for the occluded target, causing the model to fail the detection task. }\label{fig2}
\end{figure}
\section{Results}
Our experiments demonstrate that the proposed adversarial shape learning framework is a powerful and versatile tool. We systematically show that our method can: (1) generate shapes from scratch that carry sufficient semantic information to be classified as any target category by state-of-the-art models \cite{resnet,vgg,densenet,Mobilenetv2,vit,swin}; (2) serve as a novel, high-fidelity visualization tool for interpreting a network's decision-making process by identifying salient object regions; and (3) function as a generalizable adversarial paradigm, analogous to adversarial patches, that can be deployed in diverse downstream tasks such as object detection. The workflow for these experiments is illustrated in Fig. \ref{fig2}.

\subsection{Fourier shapes can embody class-specific semantic information}
\begin{figure}[!t]
	\centering
	\includegraphics[width=1.0\textwidth]{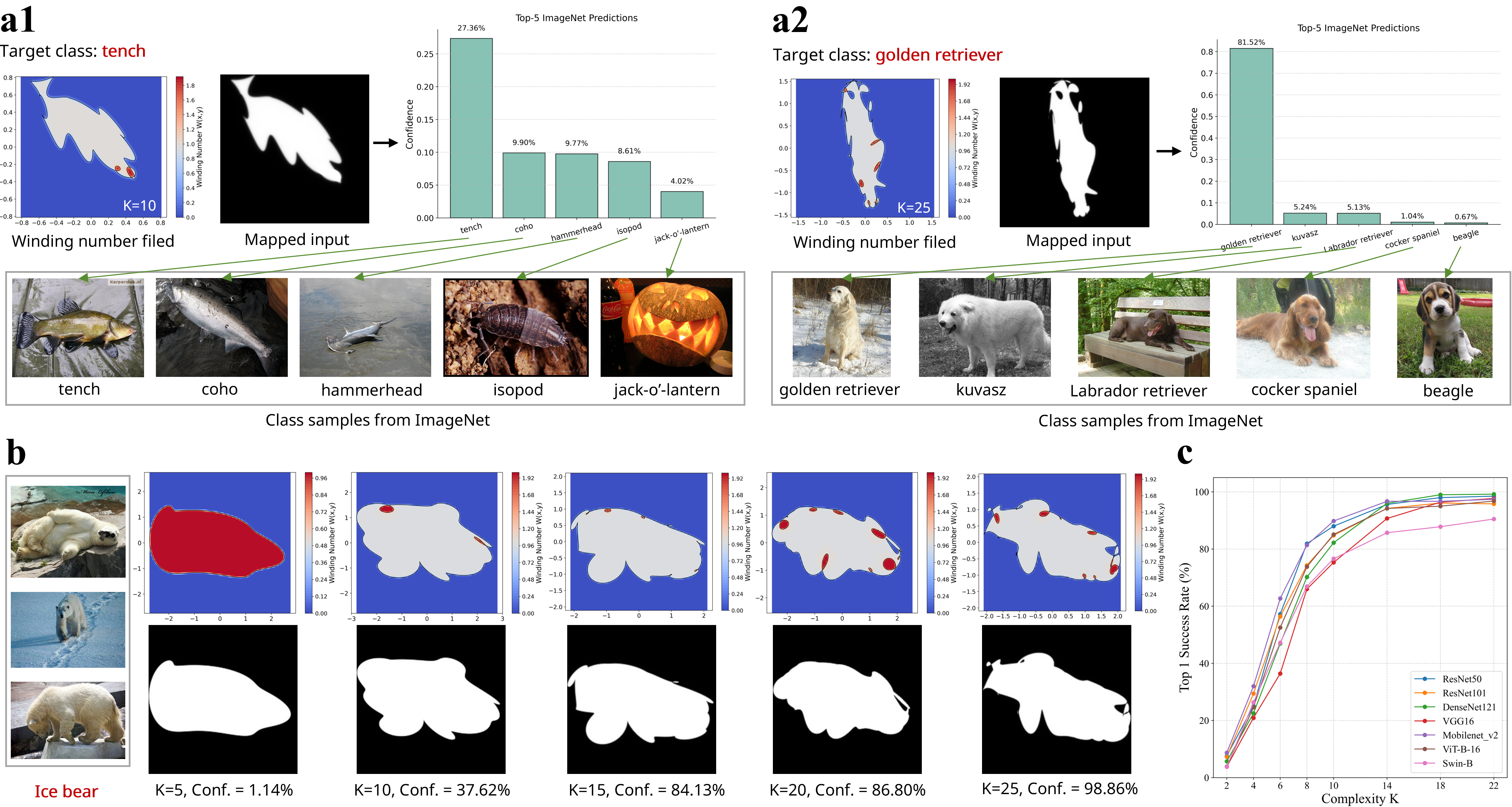}
	\caption{\textbf{Adversarial shapes generated from scratch can embody class-specific semantics}. \textbf{a}, Qualitative examples of generated shapes by the ResNet-50 model. Left, a shape generated to be classified as \emph{tench} using a complexity of $K=10$. Right, a more detailed shape generated for the \emph{golden retriever} class using $K=25$. The top-5 classification predictions and their confidence scores are listed for each shape, demonstrating high confidence for the target class and semantically logical subsequent predictions. \textbf{b}, The effect of shape complexity on classification confidence for the \emph{ice bear} class. As $K$ increases from 5 to 25, the shape incorporates more detail, and the target confidence monotonically increases from $1.14\%$ to $98.86\%$. \textbf{c}, Generalization of the learnable Fourier shape across diverse model architectures and all ImageNet classes. The plot displays the top-1 classification success rate as a function of shape complexity. Each curve represents a different model architecture. The success rate for each point is the average across all 1,000 ImageNet classes. For all models tested, the success rate consistently exceeds $90\%$ as $K$ increases beyond 20.}\label{fig3}
\end{figure}

To investigate whether shape, in complete isolation from colour and texture, can function as an effective semantic carrier for DNNs, we designed an experiment to generate class-specific shapes from scratch. We employed a targeted optimization process where the Fourier coefficients defining a shape were iteratively updated to maximize the classification confidence score for a designated ImageNet \cite{imagenet} class on a pre-trained ResNet-50 model \cite{resnet}. The input to the network was the grayscale image generated by our differentiable mapping pipeline, containing only the optimized shape against a black background. This setup allows us to directly probe the geometric priors learned by the network.

Our findings reveal that this process can successfully generate highly specific and recognizable shapes that effectively trigger the desired classification (Fig. \ref{fig3}). For instance, when targeting the \emph{tench} class, our method generates a shape that is not only classified with the highest confidence as a tench but is also intuitively recognizable to a human observer as the silhouette of a fish (Fig. \ref{fig3}a, left). This result provides strong initial evidence that the network's learned features for this class are intrinsically linked to a distinct geometric form. Notably, the network's subsequent predictions (top-5) correspond to other visually similar aquatic creatures, such as \emph{coho} and \emph{hammerhead}, suggesting that its confusion is semantically logical and rooted in shared shape characteristics, rather than being an arbitrary failure mode.

The capability of our method extends to more challenging, fine-grained categories where shape cues are subtler. When tasked with generating a \emph{golden retriever}, a category distinguished from other dog breeds by features that are often textural, the optimization required a higher shape complexity ($K=25$). The resulting shape, while more abstract to the human eye, was classified as a golden retriever with an exceptionally high confidence of $81.52\%$ (Fig. \ref{fig3}a, right). Close inspection reveals that the shape evolved to capture characteristic local details, such as the contours of the ears and paws. Again, the model's top-5 predictions were all other visually similar retriever and spaniel breeds, reinforcing the notion that our method uncovers a hierarchy of geometric features learned by the model.

A key advantage of our Fourier representation is its parametric efficiency. A standard $224\times224$ pixel-space attack requires optimizing over 50,000 parameters, whereas our shape, even with a high complexity of $K=25$, is defined by only $2K+1=51$ learnable parameters. This compactness does not sacrifice effectiveness. We found a direct and graceful correlation between the shape's complexity $K$ and the adversarial success. As $K$ increases, the shape can incorporate finer details, leading to a monotonic increase in the target class confidence score (Fig. \ref{fig3}b). This demonstrates that the model's confidence is tied to the fidelity of the geometric details present in the shape.

Finally, to confirm that this phenomenon is not specific to one model or a few object classes, we conducted a large-scale quantitative analysis. We systematically generated shapes for all 1,000 ImageNet classes across a diverse suite of seven leading architectures, including convolutional networks (ResNet-50, ResNet-101, DenseNet121 \cite{densenet}, VGG16 \cite{vgg}, MobileNetv2 \cite{Mobilenetv2}) and vision transformers (ViT-B-16 \cite{vit}, SwinTransformer-B \cite{swin}). The results show a universally consistent trend: for all tested models, the success rate of generating a shape correctly classified as the target class increases monotonically with $K$. As the number of Fourier terms increases beyond 20, most architectures achieve high success rates (above $96\%$), while the Swin-Transformer also achieves a success rate of approximately $90\%$ (Fig. \ref{fig3}c). This comprehensively demonstrates that our shape-learning framework is a general and robust method for instantiating nearly any object category conceivable by modern deep learning models, using geometry as the sole information carrier.

\subsection{Learnable shapes serve as a high-fidelity interpretability tool}
Building on the discovery that shapes can intrinsically carry class-specific semantics, we investigated whether our learnable shapes could be repurposed as a high-fidelity tool to interpret the inner workings of DNNs. A central challenge in AI interpretability is to precisely identify the minimal visual evidence a model uses to make a specific classification. Existing methods, such as gradient-based attribution maps, often produce coarse, diffuse heatmaps that highlight general areas of importance but lack precise boundaries. We therefore sought to determine if our framework could isolate these critical regions with geometric precision.

\begin{center}
	\includegraphics[width=1.0\textwidth]{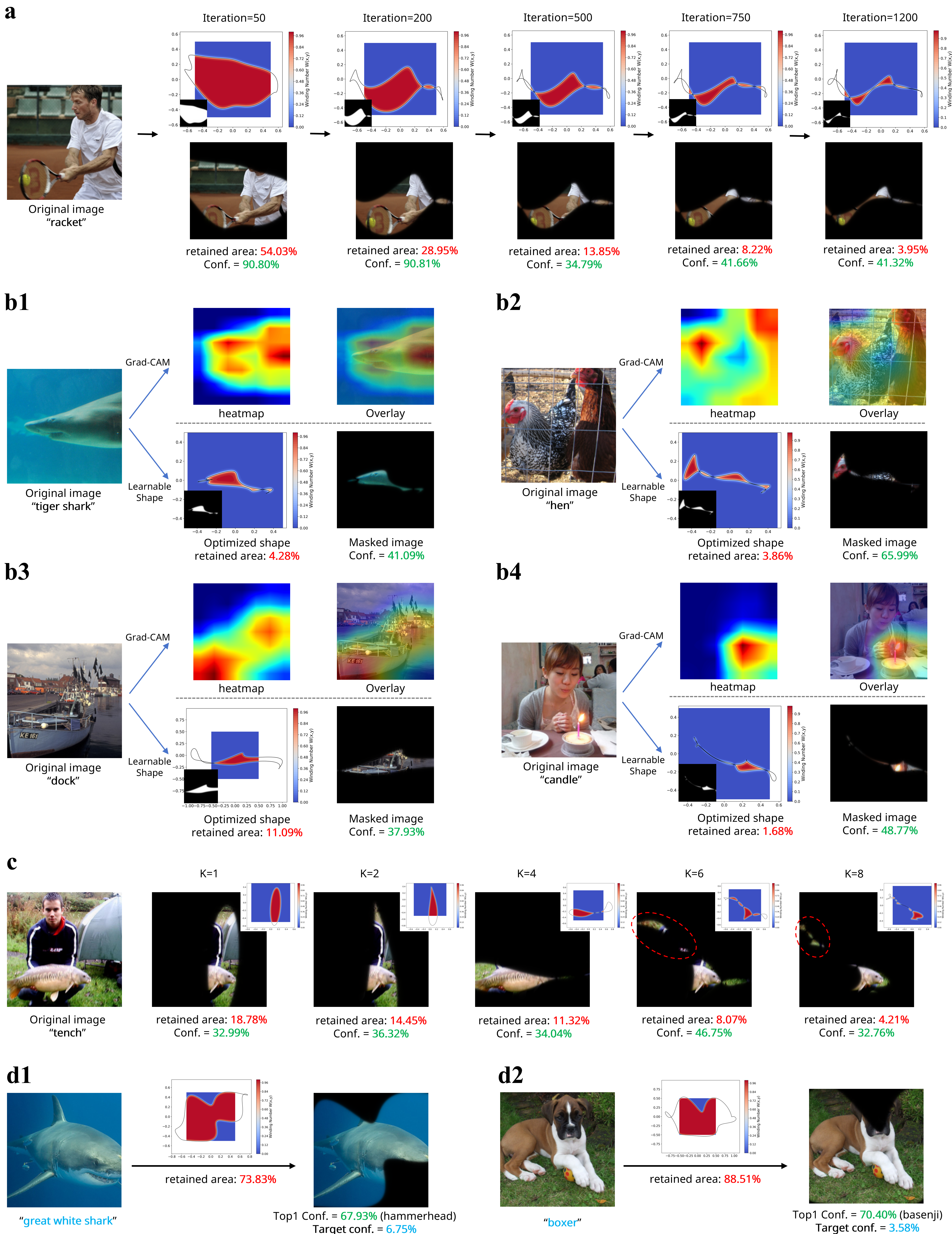}
\end{center}
\clearpage

\vspace*{\textfloatsep} 

\makeatletter  

\noindent\begin{minipage}{\textwidth}
	
	\refstepcounter{figure}
	
	\def\@captext{\textbf{Learnable Fourier shapes as a high-fidelity tool for model interpretability}. \textbf{a}, The optimization process for identifying the salient region in an image of a \emph{racket}. As the number of iterations increases from 50 to 1,200, the shape mask progressively contracts to focus on the racket head, while the retained area decreases from $54.03\%$ to $3.95\%$. Throughout this process, the masked input image is consistently classified correctly as racket with high confidence. \textbf{b}, Comparison of our learnable shape method with Grad-CAM for visualizing the salient regions for four different images. For each example (b1-b4), our method isolates a small, precise region with sharp boundaries (for example, $4.28\%$ for \emph{tiger shark} and $1.68\%$ for \emph{candle}) that is sufficient for correct top-1 classification. In contrast, the Grad-CAM heatmaps, generated from the final convolutional layer of a ResNet-50, highlight a more diffuse area. \textbf{c}, The effect of shape complexity (number of Fourier terms, $K$) on localizing the salient region for an image of a \emph{tench}. Increasing $K$ from 1 to 8 allows the shape to identify a progressively smaller critical area, reducing the retained area from $18.78\%$ to $4.21\%$. At higher complexities ($K=6$ and $K=8$), high-frequency \emph{tails} can appear in non-salient regions, as indicated by the red dashed circles. \textbf{d}, Results from the symmetric experiment, where the objective is to induce misclassification by occluding the smallest possible critical region. For the \emph{great white shark} (d1), masking only the teeth and dorsal fin (occluding $26.17\%$ of the image) results in a high-confidence misclassification to \emph{hammerhead}. For the \emph{boxer} (d2), masking the face (occluding $11.49\%$ of the image) leads to a misclassification as \emph{basenji}.}
	
	\@figurecaption{\figurename~\thefigure}{\@captext}
	
	\addcontentsline{lof}{figure}{\protect\numberline{\thefigure}\@captext}
	
	\label{fig4}
	
\end{minipage}\par 

\makeatother 

\vspace{6pt}%

To achieve this, we designed a dual-objective optimization experiment. For a given input image from the ImageNet dataset, we optimize a Fourier shape mask that is element-wise multiplied with the image. The resulting masked image, which preserves only the visual information within the shape's contour, is then fed into a ResNet-50 model. The optimization is guided by two competing objectives: maximizing the classification confidence for the image's true label while simultaneously minimizing the area of the shape mask. This process forces the shape to iteratively contract and converge upon the most compact and informative region essential for the model's decision. Furthermore, we designed a symmetric experiment to validate these findings by inverting the objectives. Specifically, we minimized the classification confidence while maximizing the mask area to identify the smallest critical region that, when occluded, guarantees misclassification.

Our findings demonstrate that the learnable shape dynamically and efficiently converges to the semantically salient regions of an object. The optimization process reveals a clear, iterative focusing effect, where an initially large and simple shape gradually refines its contour to tightly envelop the key features of the target object, such as the racket head in a tennis image (Fig. \ref{fig4}a). The final optimized shape often constitutes a remarkably small fraction of the original image area; for instance, the model can correctly classify the \emph{racket} image with high confidence even when only $3.95\%$ of the original pixels are retained.

When compared with established visualization tools like Grad-CAM \cite{Grad-cam}, our method offers substantially improved precision and interpretability (Fig. \ref{fig4}b). Grad-CAM, which relies on the gradients of the final convolutional feature maps, inherently produces low-resolution, diffuse heatmaps that are spatially coarse. In contrast, our shape-based approach generates masks with sharp, unambiguous boundaries derived directly from the winding number calculation. This allows for a much clearer delineation of the model's focus. For a \emph{candle} image, our method identified that a mere $1.68\%$ of the image, precisely covering the flame and wick, was sufficient for correct classification, providing a far more concentrated and interpretable result than the corresponding heatmap (Fig. \ref{fig4}b, bottom right). This heightened precision stems from our explicit optimization objective to minimize area, a constraint not present in attribution-based methods.

We also analyzed how the shape's complexity $K$ influences this localization task (Fig. \ref{fig4}c). We observed that shapes with higher complexity (larger $K$) are capable of identifying smaller and more intricate salient regions. This is because the increased flexibility allows the contour to \emph{carve out} non-essential areas with greater precision. However, this flexibility comes with a trade-off. At higher $K$ values (e.g., $K=8$), the optimized shape can develop self-intersections or high-frequency \emph{tails} that extend into non-salient background areas, albeit covering a negligible area. These artifacts may arise partly from the optimization dynamics, where all coefficients are updated jointly. More fundamentally, these high-frequency tails are a manifestation of the shape's own strong semantic information, as demonstrated in our first experiment. This suggests that the shape's own semantics could potentially interfere with the goal of purely isolating the image's salient features. We therefore recommend using a moderate complexity (e.g., $K=6$), as it strikes an effective balance, enabling fine-grained localization while mitigating the introduction of confounding shape-based priors.

Finally, the symmetric experiment, where we aimed to retain as much of the original image as possible while inducing misclassification, offers compelling insights into the model's failure modes (Fig. \ref{fig4}d). For an image of a \emph{great white shark}, the algorithm learned to precisely mask out the teeth and dorsal fin. Despite preserving $73.83\%$ of the image, the model's classification switched to \emph{hammerhead} with high confidence. Similarly, occluding the face of a \emph{boxer} dog was sufficient to cause a misclassification as a \emph{basenji}. These results starkly reveal the model's heavy reliance on a few local, discriminative features. Unlike human perception, which often relies on a holistic understanding of the object, the model's decision can be completely overturned by the absence of these key features, revealing a potential vulnerability in their decision-making process.

\subsection{Adversarial shapes as a generalizable attack paradigm for downstream tasks}
\begin{figure}[!t]
	\centering
	\includegraphics[width=1.0\textwidth]{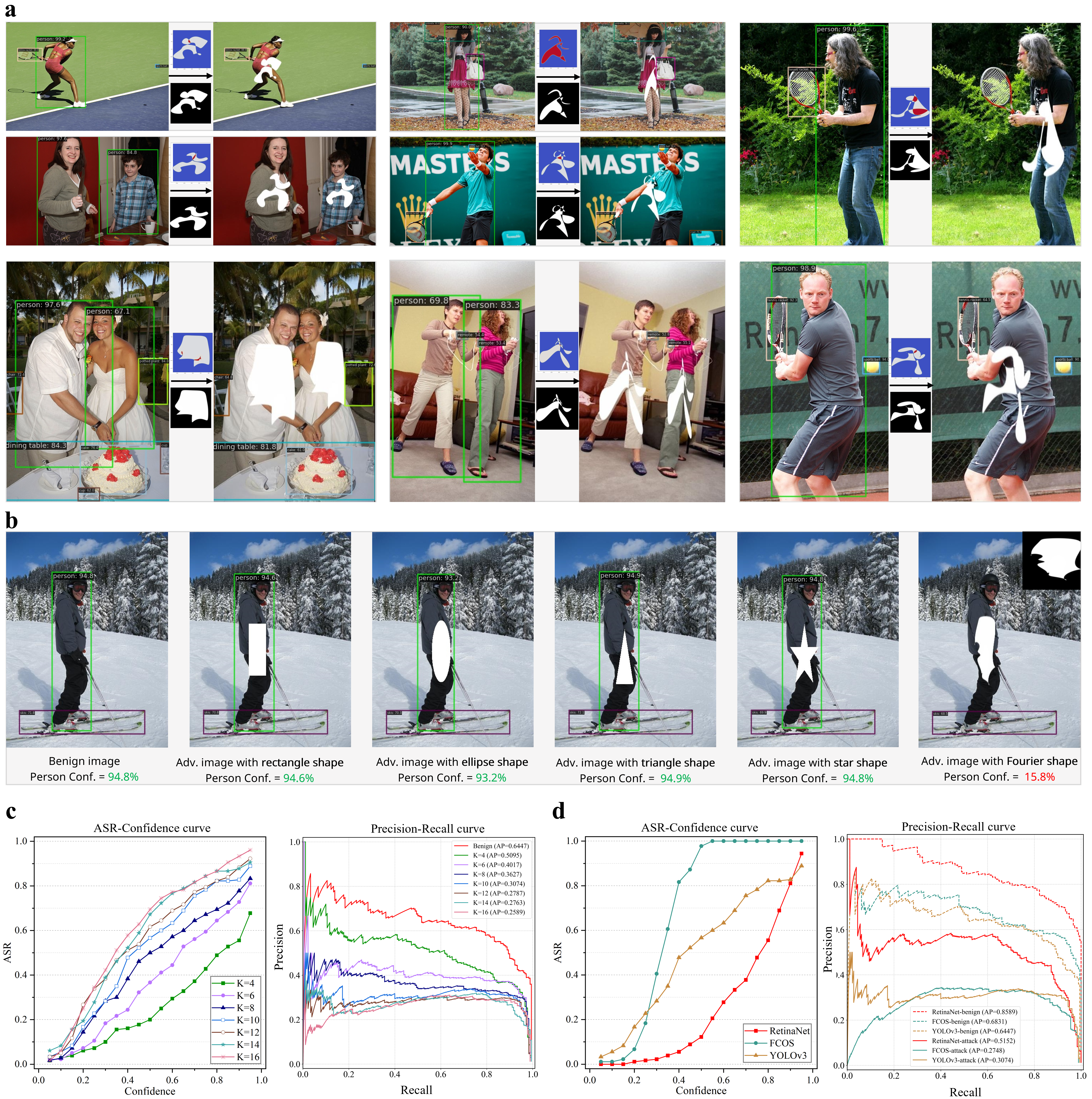}
	\caption{\textbf{Adversarial shapes as a generalizable attack paradigm for object detection}. \textbf{a}, Qualitative results of the shape attack against the YOLOv3 detector. In each pair, the left image shows the benign detection (person detected, green box) and the right image shows the attacked version. The optimized white Fourier shape ($K=10$) causes the detector to fail, and the person is no longer detected (detection confidence  $\le 0.5$). 
	\textbf{b}, Comparison of the optimized Fourier shape against simple geometric occlusions of similar area. While simple shapes (rectangle, ellipse, triangle, star) have a negligible effect on detection confidence (e.g., $93.2\%$ - $94.9\%$), the optimized shape reduces the confidence to $15.9\%$, successfully evading detection.
	\textbf{c}, Quantitative ablation on the effect of shape complexity across a set of 140 COCO images. The Attack Success Rate (ASR) vs. Confidence plot (left) shows that ASR (higher is better) increases with higher $K$. The Precision-Recall (PR) curves (right) show that the Average Precision (AP, lower is better) for the \emph{person} class decreases as $K$ increases.
	\textbf{d}, Generalization of the shape attack ($K=10$) across diverse detector architectures (YOLOv3, RetinaNet, and FCOS). The ASR-Confidence plot (left) shows the attack is effective against all models. The PR curves (right) show a significant performance degradation for all attacked models (solid lines) compared to their benign baselines (dashed lines).}
	\label{fig5}
\end{figure}

We next investigated whether the Fourier shape could be generalized to function as a new adversarial paradigm for complex, downstream vision tasks \cite{COCOdataset,yolov3,retinanet,FCOS}. This positions our method as a conceptual analogue to colour-based adversarial patches \cite{Thys_CVPRW2019,DAP_CVPR24,hu_Naturalistic,Benchmarking_tgrs}, which have proven effective in the physical world \cite{Cheng_Physical_monocular,Wang_TCSVT2025,Wang_TIP2025}. While those methods optimize the texture within a fixed, simple shape (such as a square), we invert this concept: we optimize the shape itself while keeping its internal colour fixed (e.g., solid white), thereby isolating the adversarial power of pure geometry.

To test this paradigm, we targeted the object detection, a basic task of real-world computer vision systems. We designed an experiment to make a target object, specifically a \emph{person}, \textbf{invisible} to a pre-trained YOLOv3 detector \cite{yolov3}. For a given image containing a person, our Fourier shape was rendered as a solid white patch onto the target. The patch was scaled relative to the person's bounding box (e.g., $0.6\times$ the height and width) and centred on the object. The resulting image was then fed to the detector. Our optimization objective was to minimize the objectness confidence scores for all detection proposals associated with the target, thereby causing the detector to miss the person entirely (a false negative).

The qualitative results are striking (Fig. \ref{fig5}a). In benign images, the detector robustly identifies the \emph{person} class with high confidence. After the optimized Fourier shape is applied, the person becomes invisible to the detector; the associated bounding box and confidence score disappear, even though the person remains partially visible to a human observer. This suggests the shape's optimized geometry introduces adversarial information that effectively overrides the detector's learned features for the \emph{person} category.

A critical question is whether this effect is due to the specific optimized geometry or simply to the act of occlusion. To answer this, we conducted a control experiment comparing our optimized shape to simple, non-optimized geometric shapes (e.g., a rectangle, ellipse, or star) of similar area (Fig. \ref{fig5}b). The simple shapes had a negligible impact on the detector's confidence, which remained high (e.g., $93.2\%$ to $94.9\%$). In contrast, our optimized Fourier shape decimated the confidence score to $15.9\%$, well below the typical detection threshold. This finding is crucial, as it demonstrates that the attack's potency stems not from mere occlusion, but from the specific, learned geometric contours of the shape itself.

We further quantified this effect by evaluating the attack on a set of 140 images from the COCO dataset \cite{COCOdataset}, analyzing performance as a function of shape complexity. We used two metrics: the Attack Success Rate (ASR) at different confidence thresholds, and the degradation in the model's Precision-Recall (PR) curve. A higher ASR curve indicates a more potent attack, as does a lower, more suppressed PR curve. The results clearly show that attack efficacy scales with shape complexity (Fig. \ref{fig5}c). As $K$ increases, the shape becomes more intricate, the ASR curves shift upwards, and the PR curves are pressed further downwards, indicating a greater drop in the model's Average Precision (AP).

Finally, to demonstrate the generalizability of this paradigm, we deployed the attack against three representative detectors: YOLOv3, RetinaNet \cite{retinanet}, and FCOS \cite{FCOS}. The shape attack proved universally effective, significantly degrading the performance of all three models (Fig. \ref{fig5}d). The consistent drop in the PR curves (solid lines) compared to their benign baselines (dashed lines) confirms that adversarial shapes are a robust attack vector, capable of exploiting vulnerabilities in both anchor-based (YOLOv3, RetinaNet) and anchor-free (FCOS) detectors.

These results position adversarial shapes as a viable new attack modality with significant implications for real-world robustness. Unlike texture-based patches, which are highly sensitive to colour distortion from lighting and camera sensors, a shape-based attack encodes its adversarial information in its geometry, an attribute that is more resilient to such physical-world variations. While we used a simple white patch for these experiments, this framework opens the door to hybrid attacks that could optimize both shape and colour simultaneously.

\section{Discussion}
In this work, we have established a paradigm for understanding and interacting with DNNs through the direct, holistic optimization of an object's geometry. We demonstrated the profound potential of this learnable shape framework through three distinct lines of inquiry. First, we showed that shape, in complete isolation from colour and texture, can act as a potent carrier of semantic information, capable of eliciting high-confidence, class-specific responses from well-trained models. Second, we repurposed this framework as a high-fidelity interpretability tool, capable of isolating a model's critical regions of interest with a precision and clarity that surpasses existing attribution methods. Finally, our findings establish shape as a new, generalizable modality for adversarial attacks, conceptually analogous to adversarial patches but operating in a fundamentally different domain, with broad applicability to diverse visual tasks.
The success of our approach hinges on a fully end-to-end differentiable pipeline. The use of a Fourier series provides a compact, yet powerful, parameterization for arbitrary closed contours. This abstract representation is then analytically bridged to the pixel space required by DNNs via a differentiable mapping derived from the winding number theorem. The framework is further guided by regularization constraints, inspired by signal processing principles, that ensure the generation of plausible, naturalistic shapes. Together, these components create a novel and efficient methodology for exploring the vast space of geometric forms in the context of machine perception.

The implications of this work extend beyond adversarial analysis, opening up several promising avenues for future research. First, our framework enables a more targeted data augmentation strategy. By precisely masking a model's most relied-upon features, we can compel networks to learn from a more holistic global context rather than exploiting local shortcuts, potentially leading to substantial improvements in generalization. Second, this work invites exploration into more sophisticated optimization strategies. Instead of a joint optimization of all Fourier coefficients, a staged approach could be employed. This might involve first locating a region of interest with low-frequency terms and then refining the details with high-frequency ones, perhaps using frequency-specific learning rates, which could mitigate artifacts and enhance efficiency. Finally, the principles established here can be extended to three-dimensional surfaces. By parameterizing a 3D mesh with a spherical Fourier series, one could directly optimize a 3D object to attack models operating on point clouds \cite{wen_geo2020tpami} or volumetric data \cite{tu_phy2020cvpr}. This top-down generative process would bypass many of the complex smoothness constraints required by traditional mesh manipulation methods \cite{lou_gene2024cvpr}, offering a powerful new direction for investigating and challenging the frontiers of 3D machine perception.

\section{Methods}
\subsection{Overview}
Our method introduces a novel framework for generating adversarial shapes by directly optimizing the geometric form of an object, fundamentally shifting the paradigm of adversarial attacks from the colour domain to the shape domain. Conventional adversarial attacks manipulate the input by adding subtle, often imperceptible, pixel-level perturbations to an existing image. Similarly, adversarial patches introduce a localized but fixed-shape pattern onto an object. While effective, these methods do not alter the object's intrinsic geometry. Our approach, in contrast, generates a holistic and physically realizable shape from scratch, defined by a continuous boundary, that is optimized to deceive a DNN.

The core innovation lies in creating a fully differentiable pipeline that connects a parametric representation of a closed shape to the output of a target DNN. This is achieved through three key stages. First, we model an arbitrary 2D closed shape using a Fourier series representation, which provides a compact and powerful parameterization capable of describing a vast family of complex geometries. Second, we introduce a differentiable mapping module based on the winding number theorem from complex analysis. This module analytically transforms the Fourier coefficients into a 2D rasterized image of the shape, where each pixel's value is determined by its position relative to the shape's boundary. This step is crucial as it builds a differentiable bridge between the abstract shape parameters and the pixel space that DNNs operate on. Finally, with this end-to-end differentiable pipeline, we can feed the generated shape image into the target DNN and compute the adversarial loss. The gradient of this loss is then backpropagated all the way to the Fourier coefficients, allowing us to iteratively update and \emph{grow} a shape that maximally fools the network. To ensure the generated shapes are both physically plausible and effective for attack, we introduce a set of regularization constraints based on signal processing principles, which govern the energy distribution across different frequency components of the shape.

\subsection{Shape Modeling}
To mathematically represent any arbitrary 2D closed contour in a continuous and differentiable manner, we employed a Fourier series representation. This powerful technique can approximate any periodic function as an infinite sum of sine and cosine functions. By treating the $x$ and $y$ coordinates of a shape's boundary as functions of a parameter $t$ that traverses the contour, we can define the shape in the complex plane. A shape $F(t)$ is thus represented as:
\begin{equation}
	F(t) = f(t) + i \cdot g(t) = \sum_{k=-K}^{K} c_k e^{ikt} \quad \text{for} \quad t \in [0, 2\pi]
\end{equation}
where $f(t)$ and $g(t)$ are the Cartesian coordinates of the boundary, $i$ is the imaginary unit, and $t$ is the parameter that sweeps along the contour. The shape is defined by a set of complex Fourier coefficients, $c_k=a_k+ib_k$, which are the parameters we aim to optimize. The integer $K$ determines the complexity, or degrees of freedom, of the shape.

Intuitively, each coefficient $c_k$ controls a specific geometric characteristic of the shape:
\begin{itemize}
	\item DC Offset ($c_0$): This zero-frequency term is a complex number representing the shape's centre of mass. 
	\item Fundamental Frequencies ($c_1,c_{-1}$): These terms, corresponding to $k=1$ and $k=-1$, define the fundamental elliptical or circular form of the shape. They dictate its overall scale, elongation, and orientation. A simple circle, for instance, can be defined by setting $c_1$ to a real number and all other coefficients to zero.
	\item Higher Harmonics ($c_k,|k|\geq2$): These coefficients add progressively finer details and complexity to the base ellipse. For instance, $c_2$ and $c_{-2}$ might introduce a twofold symmetry (like a peanut shape), while $c_3$ and $c_{-3}$ could add a threefold symmetry (like a cloverleaf). By combining these harmonics, we can construct an immense variety of intricate shapes.
	
\end{itemize}

The primary advantage of this representation is its compactness and differentiability. Instead of optimizing tens of thousands of pixels, we only need to optimize a small set of $2*(2K+1)$ real-valued parameters ($a_k$ and $b_k$), making the optimization process highly efficient. Crucially, the shape's coordinates $f(t)$ and $g(t)$ are analytic functions of these coefficients, which is a prerequisite for gradient-based optimization.

\subsection{Differentiable Mapping}
A key challenge is to bridge the parametric shape representation with the grid-like input required by a DNN. We need a differentiable process that can "draw" the shape onto a 2D canvas. We achieve this using a robust method derived from the winding number theorem. The winding number, $W$, quantifies how many times a closed curve travels counter-clockwise around a given point $(x_0, y_0)$. For a simple, non-self-intersecting closed curve, the winding number is 1 for any point inside the curve and 0 for any point outside. This binary property provides a perfect criterion for defining the interior of our shape.

The winding number can be calculated via the following line integral along the curve $C$:
\begin{equation}
	W(x_0, y_0) = \frac{1}{2\pi} \oint_C \frac{(x-x_0)dy - (y-y_0)dx}{(x-x_0)^2 + (y-y_0)^2}
\end{equation}
By substituting our parametric expressions $x=f(t)$, $y=g(t)$, $dx=f^{\prime}(t)dt$, and $dy=g^{\prime}(t)dt$, we can express the winding number as an integral over the parameter $t$:
\begin{equation}
	W(x_0, y_0) = \frac{1}{2\pi} \int_{0}^{2\pi} \frac{(f(t)-x_0)g'(t) - (g(t)-y_0)f'(t)}{(f(t)-x_0)^2 + (g(t)-y_0)^2} dt
\end{equation}
This integral gives the winding number for a single point $(x_0, y_0)$. To generate a full image, we evaluate this integral for every pixel coordinate $(x_p, y_p)$ in our target image grid $I$. The value of each pixel $I(p)$ is thus a function of the winding number at its location. In practice, we implement this integral in a discrete and differentiable form. By sampling $N$ points $t_j=j\cdot(2\pi/N)$ along the curve, the integral is approximated by the following differentiable sum:
\begin{equation}
	W(x_0, y_0) \approx \frac{1}{N} \sum_{j=0}^{N-1} \frac{(f(t_j)-x_0)g'(t_j) - (g(t_j)-y_0)f'(t_j)}{(f(t_j)-x_0)^2 + (g(t_j)-y_0)^2}
\end{equation}
This entire process, from the Fourier coefficients $c_k$ to the final raw image $I$, is fully differentiable. The resulting image $I$ contains pixel values that are floating-point approximations of the true winding number at each coordinate. For a simple, non-self-intersecting curve, these values will be close to 1 for the interior and close to 0 for the exterior. During optimization, the shape may self-intersect, resulting in regions where the calculated values approximate other integers (e.g., 2, -1). The key insight is that any region with a calculated value significantly deviating from zero corresponds to the shape's interior. Therefore, to create a robust mask for the DNN, we process the raw image $I$ by first taking its absolute value. This step ensures that regions approximating both 1 and -1 are treated as positive. We then clip the values to the range [0, 1]. This normalization effectively thresholds the continuous-valued winding number field, mapping all significant interior regions (where the approximate winding number's absolute value is high) towards a value of 1 and the exterior towards 0, creating a well-formed input for the network.  
The automatic differentiation engines in modern deep learning frameworks like PyTorch can therefore compute the exact gradients ${\partial}I(p)/{\partial}c_k$ for every pixel. This allows the adversarial loss, computed from the DNN's output, to flow back and directly inform the update of the shape's defining parameters.

\subsection{Regularization Constraints}
Unconstrained optimization of the Fourier coefficients can lead to shapes that are physically unrealistic or contain excessive high-frequency noise. Such shapes may be effective in simulation but are not meaningful as real-world adversarial objects. To guide the optimization towards plausible and robust shapes, we introduce two regularization terms into our loss function, based on signal energy principles.

First, we enforce \textbf{fundamental frequency dominance}. The overall structure of a natural object is typically defined by its low-frequency components. We therefore constrain the energy of the fundamental frequencies ($c_1, c_{-1}$) to be dominant over the higher harmonics. We define the sum of fundamental and harmonic amplitudes as $S_{\text{fund}} = |c_1| + |c_{-1}|$ and $S_{\text{harm}} = \sum_{|k|=2}^{K} |c_k|$, respectively. The constraint is formulated as a penalty term added to the loss if the following condition is violated:
\begin{equation} 
	S_{\text{fund}} \ge \lambda \cdot S_{\text{harm}} \label{eq:fund_dominance} 
\end{equation}
where $\lambda > 1$ is a hyperparameter that enforces the desired dominance (e.g., $\lambda=2$). This encourages the optimization to first establish a stable, low-frequency base shape before adding details.

Second, we impose an \textbf{individual higher harmonic amplitude limit}. While higher harmonics are essential for crafting the specific features that deceive the network, allowing any single harmonic to become excessively strong can create unrealistic, spiky artifacts. We therefore limit the amplitude of each individual higher harmonic coefficient to be no more than a fraction, $\gamma$, of the total fundamental amplitude:
\begin{equation} 
	|c_k| \le \gamma \cdot S_{\text{fund}} \quad \text{for all} \quad |k| \ge 2 \label{eq:harmonic_limit} 
\end{equation}
where $\gamma$ is a hyperparameter (e.g., $\gamma=0.25$). This constraint ensures that the high-frequency details serve to refine the shape rather than dominate its structure. Together, the overall regularization constraint can be formulated as:
\begin{equation}
	\mathcal{L}_{\text{reg}} = \text{ReLU}(\lambda S_{\text{harm}}-S_{\text{fund}})+\sum_{|k|=2}^{K}\text{ReLU}(|c_k|-\gamma S_{\text{fund}})
\end{equation}
The $\mathcal{L}_{\text{reg}}$ acts as a prior for plausible geometries, accelerating convergence and resulting in smoother, more robust adversarial shapes. 

\subsection{Optimization Objectives}
To formally describe the optimization process for our three main experiments, we define the following. Let $\mathbf{c} = \{c_k\}_{k=-K}^{K}$ be the set of optimizable Fourier coefficients. Let $I(\mathbf{c})$ be the normalized grayscale image generated from these coefficients. Let $\mathcal{C}(\cdot)$ be a classification network that outputs a probability distribution over classes, and let $\mathcal{D}(\cdot)$ be a detection network. 

\textbf{Experiment 1: Class-Specific Shape Generation.} To generate a shape that embodies a target class $y_{\text{target}}$, we optimize the coefficients $\mathbf{c}$ by minimizing the negative log-probability of the target class, combined with the regularization loss:
\begin{equation}
	\mathcal{L}_1(\mathbf{c}) = -\log(\mathcal{C}(I(\mathbf{c}))_{y_{\text{target}}}) + \lambda_{reg} \cdot \mathcal{L}_{\text{reg}}
	\label{eq:exp1_loss}
\end{equation}
where $\lambda_{reg}$ is a weighting hyperparameter for the regularization term.

\textbf{Experiment 2: Shape as an Interpretability Tool.} For a given natural image $\mathbf{x}_{\text{nat}}$ with true label $y_{\text{true}}$, we optimize a shape mask $I(\mathbf{c})$ that is element-wise multiplied with the image.
To identify the minimal salient region, we maximize the confidence for the true class while minimizing the mask area:
\begin{equation}
	\mathcal{L}_{2a}(\mathbf{c}) = -\log(\mathcal{C}(\mathbf{x}_{\text{nat}} \odot I(\mathbf{c}))_{y_{\text{true}}}) + \lambda_{\text{area}} \cdot \text{mean}(I(\mathbf{c})) + \lambda_{reg} \cdot \mathcal{L}_{\text{reg}}
	\label{eq:exp2a_loss}
\end{equation}
To identify the minimal region to occlude for misclassification, we minimize the confidence for the true class while maximizing the mask area:
\begin{equation}
	\mathcal{L}_{2b}(\mathbf{c}) = \log( \mathcal{C}(\mathbf{x}_{\text{nat}} \odot I(\mathbf{c}))_{y_{\text{true}}}) - \lambda_{\text{area}} \cdot \text{mean}(I(\mathbf{c})) + \lambda_{reg} \cdot \mathcal{L}_{\text{reg}}
	\label{eq:exp2b_loss}
\end{equation}
where $\lambda_{\text{area}}$ is a weighting hyperparameter for the area term.

\textbf{Experiment 3: Shape as an Adversarial Patch.} For a given image $\mathbf{x}_{\text{nat}}$ containing foreground objects specified by bounding boxes $\mathbf{B}$, we render the shape as an occlusion patch. Let $\mathcal{R}(\mathbf{x}_{\text{nat}}, I(\mathbf{c}), \mathbf{B})$ be the function that renders the shape onto the image at the specified locations. The goal is to minimize the objectness scores of all detections associated with the foreground objects occluded by the shape. Let $\{o_j\}$ be the set of these corresponding object confidence scores output by the detector $\mathcal{D}$. The loss is:
\begin{equation}
	\mathcal{L}_3(\mathbf{c}) = \sum_{j} -\log(1 - o_j) + \lambda_{reg} \cdot \mathcal{L}_{\text{reg}} \quad \text{where} \quad \{o_j\} \text{ from } \mathcal{D}(\mathcal{R}(\mathbf{x}_{\text{nat}}, I(\mathbf{c}), \mathbf{B}))
	\label{eq:exp3_loss}
\end{equation}

\end{document}